\documentclass[10pt,twocolumn,letterpaper]{article}

\usepackage{cvpr}
\usepackage{times}
\usepackage{epsfig}
\usepackage{graphicx}
\usepackage{subfigure}
\usepackage{amsmath}
\usepackage{amssymb}
\usepackage{amsthm}
\usepackage{amsfonts}
\usepackage{mathrsfs}
\usepackage{booktabs}
\usepackage{multirow, eucal}

\usepackage[vlined,ruled,linesnumbered]{algorithm2e}

\usepackage{cite}

\usepackage[pagebackref=true,breaklinks=true,letterpaper=true,colorlinks,bookmarks=false]{hyperref}

\usepackage{caption}
\def\cL{ {\cal L } }
\def\z{ {\boldsymbol z} }

\def\Z{ {\bf Z } }
\def\cS{{\cal S}}
\def\cD{{\cal D}}

 \cvprfinalcopy %

\def\cvprPaperID{15} %

\ifcvprfinal\fi

\begin{document}

\title{Fast Training of Triplet-based Deep Binary Embedding Networks\thanks{Published at Proc.\ IEEE Conference on Computer Vision and Pattern Recognition 2016.
Code can be downloaded at \url{http://bit.ly/2asfI14}}.}

\author{Bohan Zhuang, Guosheng Lin, Chunhua Shen\thanks{C. Shen is the corresponding author; e-mail: (\url{chunhua.shen@adelaide.edu.au}).}, Ian Reid\\
  The University of Adelaide; and Australian Centre for Robotic Vision
  }

	\maketitle
	\thispagestyle{empty}

	\begin{abstract}

		In this paper, we aim to learn a mapping (or embedding) from images to a compact binary space in which Hamming distances correspond to a ranking measure for the image retrieval task.
       We make use of a triplet loss because this has been shown to be most effective for ranking problems.
         However, training in previous works can be prohibitively expensive due to the fact that optimization is directly performed on the triplet space, where the number of possible triplets for training is cubic in the number of training examples.
         To address this issue, we propose to formulate high-order binary codes learning as a multi-label classification problem by explicitly separating learning into two interleaved stages.
         To solve the first stage, we design a large-scale high-order binary codes inference algorithm to reduce the high-order objective to a standard binary quadratic problem such that graph cuts can be used to efficiently infer the binary codes which serve as the labels of each training datum.
         In the second stage we propose to map the original image to compact binary codes via carefully designed deep convolutional neural networks (CNNs) and the hashing function fitting can be solved by training binary CNN classifiers.
          An incremental/interleaved optimization strategy is proffered to ensure that these two steps are interactive with each other during training for better accuracy.
          We conduct experiments on several  benchmark datasets, which demonstrate both improved training time (by as much as two orders of magnitude) as well as producing state-of-the-art hashing for various retrieval tasks.

	\end{abstract}

\tableofcontents
\clearpage

	\section{Introduction}

	\begin{figure}[tbp]
		\centering
		\resizebox{\linewidth}{!}
		{
			\includegraphics[width=0.15\linewidth, height=0.1\linewidth]{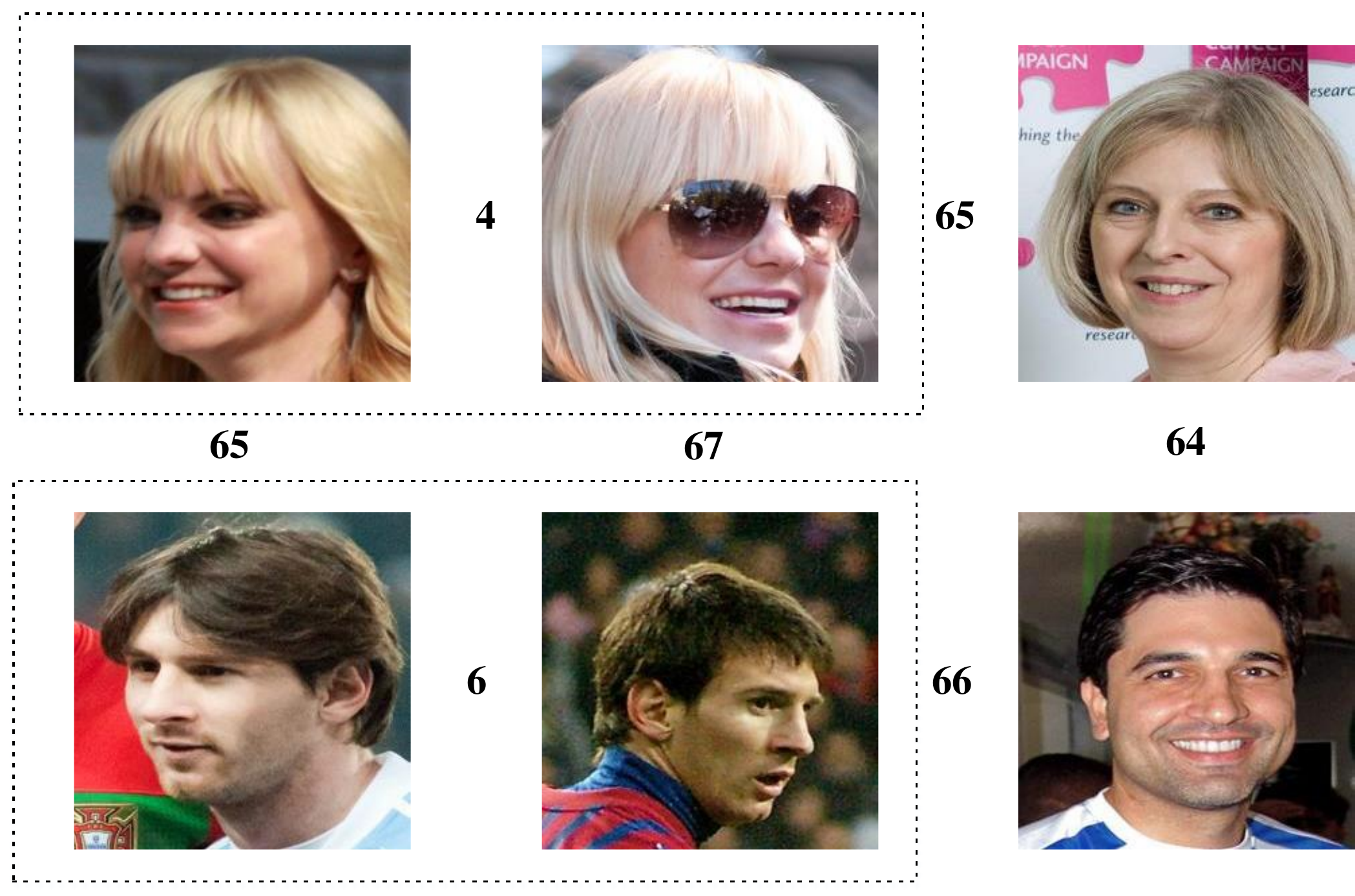}
		}
		\caption{
			The  Hamming distances calculated using the
			proposed hashing framework between pairs of faces.
			Each row represents a triplet of samples and the face pairs enclosed by a rectangle are from
			the same identity.
			Here each face image is represented by a 128-dimensional binary codes vector.
			We can see that a threshold of about 63 can correctly classify same-identity and different-identity
			pairs of faces.
		}
		\label{fig:triplet}
	\end{figure}

	With the rapid development of big data,
	large-scale nearest neighbor search with binary hash codes has attracted much more attention.
	Hashing methods aim to map the original features to compact binary codes
	that are able to preserve the {\em semantic} structure of the original features in the Hamming space.
	Compact binary codes are extremely suitable for efficient data storage and fast search.
	A few  hashing methods in the literature incorporate the triplet ranking
	loss to
	learn codes that preserve relative similarity relations~\cite{norouzi2012hamming, lai2015simultaneous, zhao2015deep, zhang2015bit,li2013learning}.
	In these works usually a triplet ranking loss is defined,
	followed by solving an expensive optimization problem.
	For instance, Lai \etal~\cite{lai2015simultaneous} and Zhao \etal~\cite{zhao2015deep} map original
	features into binary codes via deep convolutional neural networks (CNNs).  Both use a
	triplet ranking loss designed to preserve relative similarities, with the key difference being in the exact form of the loss function used.
	Similarly, FaceNet \cite{schroff2015facenet} uses the triplet loss to learn a real-valued compact embedding of faces.
	All these methods suffer from huge training complexity, because they directly train the CNNs using the triplets, the number of which scales cubically with the number of images in the training set.
	For example, the training of FaceNet \cite{schroff2015facenet} took a few months
	on Google's computer clusters.  Other work like \cite{wang2014learning}  simply subsamples a small subset
	to reduce the computation complexity.

	To address this issue, we employ a collaborative two-step approach, originally proposed in \cite{lin2013general},
	to avoid directly learning hash functions based on the triplet ranking loss.
	This two-step approach enables us to
	convert triplet-based hashing into an efficient combination of solving binary quadratic programs and learning conventional CNN classifiers.  Hence, we don't need to directly optimize the loss function with huge number of triplets to learn deep hash functions. The result
	is an algorithm with computational complexity that is orders of magnitude lower than existing work such as \cite{zhao2015deep, schroff2015facenet}, but without sacrificing accuracy.
	The two-step approach to hashing advocated by \cite{lin2014fast, lin2013general} uses
	decision trees as hash functions in combination with the design of efficient binary code inference methods.
	The main difference of our work
	is as follows.
	The work in \cite{lin2014fast, lin2013general}  only preserves the {\em pairwise} similarity relations which do not directly
	encode relative semantic similarity relationships that are important for ranking-based tasks.
	In contrast, we use a triplet-based ranking loss to preserve relative semantic relationships.
	However it is not trivial to extend the first step (binary code inference) in  \cite{lin2014fast} to
	triplet-based loss functions.  The formulated binary quadratic problem (BQP) in \cite{lin2014fast}
	can be viewed as a pairwise Markov random field (MRF) inference problem, while in our case we need to solve large-scale {\em high-order} MRF inference.
	We here propose an efficient high-order binary code inference algorithm, in which we equivalently convert the binary high-order inference into the second-order binary quadratic problem, and graph cuts based block search method can be applied.
	In the second step of hash function learning, the work of \cite{lin2014fast, lin2013general} relies on training classifiers
	such as linear SVM or decision trees on handcrafted
	features. We instead fit deep CNNs with incremental optimization to simultaneously learn feature representations and hash codes.

	Our contributions are summarized as follows.
	\begin{itemize}
		\item{
			To address the issue of prohibitively high computational complexity in triplet-based binary code learning,
			we propose a new efficient and flexible  framework for
			interactively inferring binary codes and learning the deep hash functions, using a triplet-based loss function.
			We show how to convert the high-order loss introduced by the triplets into a binary quadratic problem that can be optimized efficiently in the manner of \cite{lin2014fast}, using block-coordinate descent with graph-cuts.  To learn the mapping from images to hash codes, we design deep CNNs capable of preserving their semantic ranking information of the data.

		}

		\item{
		We
			propose a novel incremental group-wise training approach, that interleaves finding groups of bits of the hash codes, with learning the hash functions.
			We show experimentally that this approach improves the quality of hash functions while retaining the advantage of efficient training.

		}

		\item{
			We demonstrate that our method outperforms many existing state-of-the-art hashing methods on
			several benchmark datasets by a large margin.
			We also demonstrate our hashing method in the context of a face search/retrieval
			system.  We achieve the best reported results on face search under the IJB-A protocol.
		}

	\end{itemize}

	\subsection{Related work}
	Hashing methods may  be roughly categorized into data-dependent and data-independent schemes.
	Data-independent methods~\cite{gionis1999similarity, kulis2009kernelized, Jiang_2015_CVPR} focus on using
	random projections to construct random hash functions.
	The canonical example is the locality-sensitive hashing (LSH)~\cite{gionis1999similarity},
	which offers guarantees that metric similarity is preserved for sufficiently long codes based on random projections.
	Recent research focuses have been shifted to data-dependent methods, which learn hash functions in a
	either unsupervised, semi-supervised, or supervised learning fashion.
	Unsupervised hashing methods~\cite{Carreira-Perpinan_2015_CVPR, gong2013iterative, liu2011hashing, weiss2009spectral, weiss2012multidimensional, shen2013inductive} try to map the original features into hamming space while preserving similarity relations between the original features using unlabelled data.
	Supervised methods~\cite{erin2015deep, shen2015supervised, kulis2009learning, liu2012supervised, li2013learning} use labelled training data for the similarity relations, aiming to preserve the ``ground truth'' similarity in the hash codes.
	Semi-supervised hashing methods incorporate ground-truth similarity information for the subset of the training data for which it is available, but also use unlaballed data.
	Our proposed method belongs to the supervised hashing framework.

	Recently hashing using deep learning  has shown great promise.
	The authors of \cite{zhao2015deep,lai2015simultaneous} learn hash bits
	such that multilevel semantic similarities  are kept, taking raw pixels as input
	and training a deep CNN. This has the effect of simultaneously learning an image feature representation (in the early layers of the network) and the hash bits, which are obtained by thresholding the outputs of the last network layer, or \emph{hash layer} at 0.5.
	Note that these methods suffer from huge computation complexity introduced by the triplet ranking loss for hashing.
	In contrast, our proposed method is much more efficient in training, as shown in our experiments.

	\section{The proposed approach}

	Our general problem formulation is as follows. Let %
	$\cD = \{ (i,j,k)\,|\,s({{\bf{x}}_i},{{\bf{x}}_j}) > s({{\bf{x}}_i},{{\bf{x}}_k})\}$ be a set of training triplet samples,
	in which $s(\cdot,\cdot)$ is some semantic similarity measures, ${\bf{x}}_i$ is the $i$-th training sample and ${\bf{x}}_i$ is semantically more similar to ${\bf{x}}_j  $ than to ${\bf{x}}_k $.
	Let $h({{\bf{x}}}) \in {\{  - 1,1\} ^q}$ be the $q$-bit hash codes of image ${{\bf{x}}}$.
	We simplify the notation by rewriting $h({{\bf{x}}_i})$, $h({\bf{x}}_j )$ and $h({\bf{x}}_k )$
	using $\z_i$, $\z_j$ and $\z_k $, respectively.
	Our goal is to learn embedding hash functions $h(\cdot) $
	to preserve the relative similarity ranking order for the images after being mapped into the binary Hamming space.
	For that purpose, we define a general form of loss functions:
	\begin{equation} \label{eq:general}
		\mathop {\min }\limits_\Z  \sum\limits_{(i,j,k) \in \cD} { \cL(  \z_i , \z_j , \z_k } ),
		{\rm{s.t.}}\,\,\Z \in {\{  - 1,1\} ^{q \times n}}.
	\end{equation}
	Here $\Z$ is the matrix that collects  binary codes for all the $ n $  data points and $q$ is the bit length.
	$ \cL $ is a triplet loss function.

	Unlike approaches such as \cite{zhao2015deep}, our method shares the advantage of \cite{lin2013general} that we are not tied to a specific form of the loss.  One typical example of losses that could be used include the \textit{Hinge ranking loss}:
	\begin{equation} \label{eq:hinge}
		\begin{array}{*{20}{l}}
			\cL(\z_i, \z_j, \z_k) =  {\max (0,\,q/2 - ({d_H}} ({\z_i},\z_j ) - {d_H}({\z_i},\z_k )).
		\end{array}
	\end{equation}
	Here ${d_H}(\cdot,\cdot)$ is the Hamming distance.

	\begin{figure*}[tbp]
		\centering
		\resizebox{\linewidth}{!}
		{
			\includegraphics[width=0.75\linewidth]{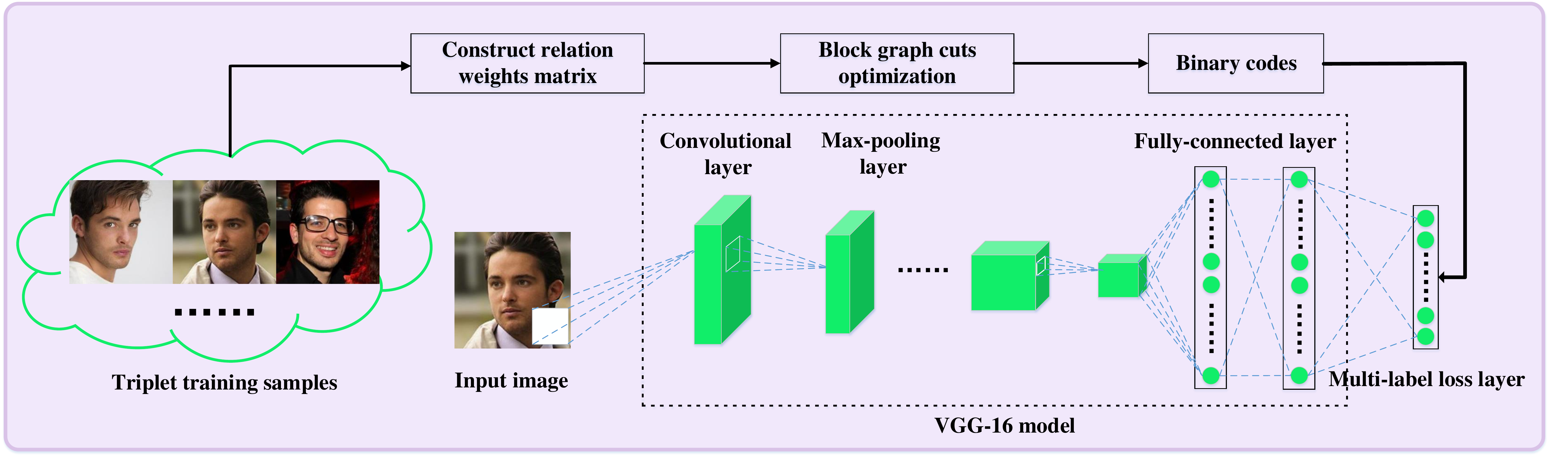}
		}
		\caption{Overview of the proposed hashing framework for training one group of binary codes.
			The framework includes two steps: binary code inference  and  hash function learning with multi-label CNNs.
			The inferred binary codes are needed by the multi-label layer of the deep hash functions.
			The CNN structure of the first a few layers is same as the VGG-16 network.
		}
		\label{fig:framework}
	\end{figure*}

	We propose an approach to learning binary hash codes that proceeds in two stages.  The first stage uses the labelled training data to infer a set of binary codes in which the hamming distance between codes preserves the semantic ranking between triplets of data.  The second stage uses deep CNNs to learn the mapping from images to the binary code space (i.e. to learn the hash functions).  A similar two-stage approach was advocated in \cite{lin2014fast}, but that work used only pairwise data, and used boosted decision trees rather than deep CNNs to learn the hash functions.

	There are various difficulties associated with direct application of triplet losses, and of CNNs to the problem.
	First, the binary code learning stage requires optimization of Eq.~(\ref{eq:general}) which is in general NP-hard.
	In Sec.~\ref{sec:inference}, we describe how to infer binary codes with triplet ranking loss by reducing the problem to a binary quadratic program.  The use of triplets considerably complicates this process and so this is one of our significant contributions in this paper.  Second, while the two-stage approach gains significantly in training time, it has the disadvantage that the learning of the codes and the hash functions do not interact and therefore cannot be mutually beneficial.  We propose a method to interleave the code and hash function learning into groups of bits, a process that retains much of the training efficiency, but improves the quality of the codes and hash functions considerably.  We explain our use of CNNs and this interleaved and incremental learning in Sec. \ref{sec:deep} below.

	\section{Inference for binary codes with triplet ranking loss} \label{sec:inference}
	Since simultaneously infer multiple bits are intractable in inference task, inspired by the work of \cite{lin2014fast}, we sequentially solve for one bit at a time conditioning on previous bits.
	When solving for the $r$-th bit, the previous $r-1 $ bits are fixed.
	The binary inference problem becomes minimization of the following objective:
	\begin{align}\label{eq:general2}
		\sum\limits_{(i,j,k) \in \cD}& {\cL({z_{r,i}}} ,{z_{r,j}},{z_{r,k}};\z_i^{(r - 1)},\z_j^{(r - 1)},\z_k^{(r - 1)}), \notag\\
		&= \sum\limits_{(i,j,k) \in \cD} {{\ell _r}} ({z_{r,i}},{z_{r,j}},{z_{r,k}}),
	\end{align}
	where ${\ell _r}$ is the loss function output of the $r$-th bit conditioned on the previous bits.
	$z_{r,i}$ is the binary code of the $i$-th data point and the $r$-th bit,
	${\z}_i^{(r - 1)}$ is the binary code vector of the previous $ r-1$ bits for the $i$-th data point.

	\subsection{Solving high-order binary inference problem}

	Directly optimizing the loss function which involves high-order relations (more than pairwise relations)
	in Eq.~(\ref{eq:general2}) is difficult since the optimization involves an extremely large number of triplets, and so can be computationally intractable.
	To address this problem, we show here how to convert the high-order inference task to a second-order problem which is much more feasible to be optimized.
	The key ``special properties'' of the binary space that we rely on are: (i) the possibility of enumerating all possible inputs (there are $2^3=8$); (ii) the symmetry of the hamming distance $d(.,.)$.
	Based on this, the triplet loss can be decomposed into a set of second-order combinations as:
	\begin{equation}\label{eq:decompose}
		\begin{array}{l}
			{\ell _r}({z_{r,i}},{z_{r,j}},{z_{r,k}}) = {\alpha _{ii}}{z_{r,i}}{z_{r,i}} + {\alpha _{ij}}{z_{r,i}}{z_{r,j}} + {\alpha _{ik}}{z_{r,i}}{z_{r,k}}\\
			+ {\alpha _{ji}}{z_{r,j}}{z_{r,i}} + {\alpha _{jj}}{z_{r,j}}{z_{r,j}} + {\alpha _{jk}}{z_{r,j}}{z_{r,k}} + {\alpha _{ki}}{z_{r,k}}{z_{r,i}}\\
			+ {\alpha _{kj}}{z_{r,k}}{z_{r,j}} + {\alpha _{kk}}{z_{r,k}}{z_{r,k}},
		\end{array}
	\end{equation}
	where ${\alpha_{..}}$ are the coefficients of the corresponding second-order combinations. Then we will show that there exists a solution for $\boldsymbol{\alpha}$ to make it a valid decomposition.
	Here we ignore the redundant terms in Eq. (\ref{eq:decompose}), hence it can be rewritten as
	\begin{align} \label{eq:simple_decompose}
		{\ell _r}({z_{r,i}},{z_{r,j}},{z_{r,k}}) &= {\alpha _{ii}}{z_{r,i}}{z_{r,i}} + {\alpha _{ij}}{z_{r,i}}{z_{r,j}} \notag  \\
		&+ {\alpha _{ik}}{z_{r,i}}{z_{r,k}}+ {\alpha _{jk}}{z_{r,j}}{z_{r,k}}= {\boldsymbol{\alpha}}^{{\rm T}}{\bf{v}},
	\end{align}
	\vspace{-5mm}
	\begin{align}
	\text{where, \;\;}	{\boldsymbol{\alpha}} &= [{\alpha_{ii}},{\alpha_{ij}},{\alpha_{ik}},{\alpha_{jk}}], \notag \\
		{\bf{v}} &= [{z_{r,i}}{z_{r,i}},{z_{r,i}}{z_{r,j}},{z_{r,i}}{z_{r,k}},{z_{r,j}}{z_{r,k}}]. \notag
	\end{align}
	$\ell_r$ has 8 possible input combinations for $({z_{r,i}},{z_{r,j}},{z_{r,k}})$  (or equivalently $\bf{v}$ has 8 possible value combinations), leading to 8 constraints of the form of (\ref{eq:simple_decompose}).  Because the loss is defined on Hamming distance/affinity, changing the sign of every input leads to identical value of the loss, thus some of these combinations lead to redundant constraints.  Eliminating all these redundant combinations leaves only four independent equations (\ref{eq:simple_decompose}).  Stacking these so that each ${\bf v}$ forms a row of a matrix yields the follow set of equations:
	\begin{equation}\label{eq:equation}
		\left[ \begin{matrix}
			1 & 1 & 1 & 1\\
			1 &	1 & -1 & -1 \\
			1 & -1 &  1 & -1 \\
			1 & -1 & -1 &  1
		\end{matrix} \right] {\boldsymbol{\alpha}} =  \left[ \begin{matrix} {\ell _r}(1,1, 1) \\ {\ell _r}(1,1, - 1) \\ {\ell _r}(1, - 1,1) \\ {\ell _r}(1, - 1, - 1) \end{matrix} \right].
	\end{equation}
	which can be easily inverted to yield the unique solution of $\boldsymbol{\alpha}$.
	This shows that for a given triplet loss function, we can decompose it into a set of pairwise terms for each triplet.

	We now seek a solution for ${\z_{(r)}}$ -- the $r^\text{th}$ bit of the code for every data point -- that optimizes the triplet relations.  Because the triplet relations are now encoded as pairwise relations, we can solve for ${\z_{(r)}}$ as follows.  We define ${\bf{W}} \in {R^{n \times n}}$ as a weight matrix in which $(i,j)$-th element of $\bf{W}$, ${w_{ij}}$, represents a relation weight between the $i$-th and $j$-th training points.
	Specifically, each element of $\bf{W}$ is computed as
	\begin{equation}
		{w_{ij}} = \sum\limits_{\forall (i,j)} {{\alpha_{ij}}},
	\end{equation}
	where ${\alpha_{ij}}$ are the coefficients corresponding to the pair $(i,j)$.  There will be one such $\alpha_{ij}$ for every triplet in which data points ${\bf x}_i$ and ${\bf x}_j$ appear.

	The triplet optimization problem in Eq.~(\ref{eq:general2}) can now be equivalently formulated as
	\begin{equation} \label{eq:BQP}
		\mathop {\min }\limits_{{\z_{(r)}} \in {{\{  - 1,1\} }^n}} \z_{(r)}^{\rm T}{\bf{W}}{\z_{(r)}}.
	\end{equation}
	Note that the coefficients matrix $\bf{W}$ is sparse and symmetric, therefore Eq.~(\ref{eq:BQP}) is a standard binary quadratic problem.
	Although we have now shown how to convert the third-order objective in Eq. (\ref{eq:general2}) into a  second-order formulation amenable to BQP, a further issue remains: the quadratic objective above contains non-submodular terms, and is therefore difficult to optimize.
	\begin{algorithm}[!t]
		\KwIn{Training images: $\{ {{\bf{x}}_1},...{{\bf{x}}_n}\}$; Relation weights matrix: $\textbf{W}$.}
		\KwOut{Sub-modular blocks: \{$\cS_1$,$\cS _2$,...\}. }
		${\cal U} \leftarrow \{ {{\bf{x}}_1},...,{{\bf{x}}_n}\}$; $t = 0$\;
		\While{${\cal U}  \ne \emptyset$}
		{
			$t=t+1$; $\cS_t  \leftarrow \emptyset$; choose an arbitrary ${{\bf{x}}_i}$ from ${\cal U}$\;
			Let ${\cal H}$ be ${\cal U} \cup \{{\bf{x}}_j | {w_{ij}} < 0  \}$

			\For{each ${{\bf{x}}_j}$ in ${\cal H}$}
			{ \If { ${w_{jk}} \le 0$ for $k=1,2,...,|\cS_t| $ }
				{Add ${{\bf{x}}_j}$ to $\cS_t$;
					If ${{\bf{x}}_j} \in {\cal U}$, remove it;
				}
			}
		}
		\caption{Greedy method for constructing blocks}
		\label{algo:construct_blocks}
	\end{algorithm}

	To address this, we follow the proposal in~\cite{lin2014fast}.  This proceeds by creating a set of sub-problems (or ``blocks'') each involving a subset of the variables $\z_{(r)}$ in which the pairwise relations are all sub-modular.  The sub-problems are then solved in turn, treating the variables that are not involved in the current block as constants.
	The inference problem for one block is written as
	\begin{equation}
	\mathop {\min }\limits_{{\z_r} \in {{\{  - 1,1\} }^n}} \sum\limits_{i \in \cS} {{u_i}} {z_{r,i}} + \sum\limits_{i \in \cS} {\sum\limits_{j \in \cS} {{v_{ij}}} } {z_{r,i}}{z_{r,j}},
	\end{equation}
	\vspace{-5mm}
	\begin{align} \label{eq:potential}
	\text{where,\;\;} {u_i} = 2\sum\limits_{j \notin \cS} {{w_{ij}}} {z_{r,j}}, \,\,   {v_{ij}} = {w_{ij}}, \notag
	\end{align}
	and $\cS$ is the block to be optimized.
	Since the above inference problem for one block is sub-modular, we can solve it efficiently using graph cuts.

	Algorithm~(\ref{algo:construct_blocks}) details how the blocks are defined.  It is subtly different from \cite{lin2014fast}; because we are using a triplet loss, the criterion for inclusion in a block is to ensure $w_{ij}<0$ for each pair ${\bf x}_i, {\bf x}_j$ in the block, which guarantees sub-modularity for all pairs.

	\begin{algorithm}[!t]

		\KwIn{Training images: $\{ {{\bf{x}}_1},...{{\bf{x}}_n}\}$; Relation map: $\textbf{M}$; group length: $a$; number of groups: $b$.}
		\KwOut{The deep hash functions: $h(\cdot)$. }
		\For{$i = 1,...b$ }
		{\For{ $j = 1,...a$}
			{
				Solve linear equations to construct the relation weight matrix $\bf{W}$\;
				Apply Block Graph-Cut algorithm~\cite{lin2014fast} to solve $((i - 1) \times a  + j)$-th bit hash codes\;
			}
			Learn the deep hash functions ${h(\cdot)}$ based on $i \times a$ bits hash codes\;
			Simultaneously update $i \times a$ bits hash codes by the output of ${h(\cdot)}$.
		}
		\caption{Two-step approach for learning deep binary embedding networks}
		\label{algo:overview}
	\end{algorithm}

	\subsection{Loss function}

	The discussion above provides a general framework for learning the binary codes using a triplet loss, but is agnostic to the exact form of the loss.  In the experiments reported in this paper, we use ${\ell _r}$ as the triplet-based hinge loss function defined in Eq. (\ref{eq:hinge}):
	\vspace{-2mm}
	\begin{equation}\label{eq:specific}
		\ell_r(...) =  {\max (0,\,\,r/2 - \Delta d_H^{(r - 1)}}  - \Delta d_H^r),
	\end{equation}
	where,
	\begin{align}
		\Delta d_H^{(r - 1)}  &= {d_H}({\z}_i^{(r - 1)},\z_j^{(r - 1)}) - {d_H}(\z_i^{(r -1)},\z_k^{(r - 1)}), \notag \\
		\Delta d_H^r &= {d_H}({z_{r,i}},{z_{r,j}}) - {d_H}({z_{r,i}},{z_{r,k}}). \notag
	\end{align}

\section{Deep hash functions learning} \label{sec:deep}

Our general scheme now requires that we learn hash functions $h(.)$ that map from data points ${\bf x}_i$ to binary codes.  We propose to do this using deep CNNs because they have repeatedly been shown to be very effective for similar tasks.  The straightforward approach is then to use the training samples, and their known codes as the labelled training set for a standard CNN.  As we have noted this two-stage approach yields significant training time gains.

However a major disadvantage is that because the binary codes are determined independently of the hash functions, and the hash functions have no possibility to influence the choice of binary codes.  Ideally these stages would interact so that the choice of binary hash codes is influenced not only by the ground-truth relative similarity relations but also by how hard the training points are.

To address this, we propose an interleaved process where we infer a group of bits within a code, followed by learning suitable hash functions for that set of bits and its predecessors, followed in turn by inference of the next group of bits, and so on.  This provides a compromise between independently learning the codes and hash functions, and a more end-to-end -- but very expensive -- approach such as \cite{lai2015simultaneous}.

\subsection{Incremental optimization} \label{sec:incremental}
Our key idea here is to optimize the hashing framework in an incremental group-wise manner.
More specifically, we assume there are $b$ groups of bits and each group has $a$ bits (e.g., for 64-bit codes we may break this into 8 groups of 8 bits each).
For convenience, we shall refer to inference of the $p$-th group binary codes followed by learning the deep hash functions, as the ``$p$-th training stage''.
In the $p$-th training stage, we first infer the $a$ bits of the $p$-th group one bit at a time (as described in Sec. \ref{sec:inference}) and then train the network parameters $\theta$ so that it minimizes the cross-entropy loss:
\begin{equation} \label{eq:cross_entropy}
- \sum\limits_{\rho  = 1}^r {\sum\limits_{i = 1}^n {[\delta ({z_{\rho ,i}} = 1)\log {{z'}_{\rho ,i}}} }  + \delta ({z_{\rho ,i}} =  - 1)\log (1 - {z'_{\rho ,i}})],
\end{equation}
where $\delta (\cdot)$ is the indication function.
Here at the $p$-th stage we are targetting the first $r=pa$ bits of the code; ${z'_{\rho,i}}$ is the $\rho$-th output of the last sigmoid layer for the $i$-th training sample; ${{z_{\rho,i}}}$ is the corresponding bit of the binary code obtained from the inference step which serves as the target label of the multi-label classification problem above. Note that in the $p$-th training stage, the bits from all $p$ groups are used to guide the learning of the deep hash functions.

Having completed training the hash functions, we then update the binary codes for all $p$ groups by the output of the learned hash functions. The effect of this is to ensure that the error in the learned hash functions will influence the inference of the next group of hash bits.

This incremental training approach adaptively regulates the binary codes according to both the fitting capability of the deep hash functions and the properties of the training data, steadily improving the quality of hash codes and the final performance.
Finally, we summarize our hashing framework in Algorithm \ref{algo:overview}.

\subsection{Network architecture}
The network of learning deep hash functions
consists of multiple convolutional, pooling, and fully connected layers (we follow the VGG-16 model),
and a multi-label  loss layer for  multi-label classification.

We use the pre-trained  VGG-16 \cite{simonyan2014very} model for initialization,
which is trained on the large-scale ImageNet dataset.
The multiple convolution-pooling and fully connected layers are used to capture mid-level image representations.
The intermediate output of the last fully connected layer are mapped to a multi-label layer as the feature representation.
Then neurons in the multi-label layer are activated by a sigmoid function so that the activations are approximated to $[0,1]$, followed by the cross-entropy loss of Eq. (\ref{eq:cross_entropy}) for multi-label classification.
\section{Experiments}
\textbf{Experimental settings}
We test the proposed hashing method on two multi-class datasets, one multi-label dataset and one face retrieval dataset.
For multi-class datasets, we use the MIT Indoor dataset~\cite{quattoni2009recognizing} and CIFAR-10 dataset~\cite{Krizhevsky09learningmultiple}.
The  MIT Indoor dataset contains 67 indoor scene categories, and 6,700 images for evaluation.
CIFAR-10 contains 60,000 small images in 10 classes.
For multilevel similarity measurement, we test our method on the multi-label dataset NUS-WIDE~\cite{chua2009nus}.
The NUS-WIDE dataset is a  large database containing 269,648 images annotated with 81 concepts.
We compare the search accuracies with four recent state-of-the-art state-of-the-art hashing methods,
including  SFHC \cite{lai2015simultaneous} (the recent deep CNNs method),
FSH~\cite{lin2014fast} (two-step hashing approach using decision trees),
KSH \cite{liu2012supervised} and ITQ \cite{gong2013iterative}.

For fair comparison, we evaluate the compared hashing methods FSH, KSH and ITQ on the features obtained from
the activations of the last hidden layer of the VGG-16 model pre-trained on the ImageNet ILSVRC-2012 dataset~\cite{ILSVRC15}.
We find that using deep CNN features in general improve the performance for these three hashing methods, compared with what was
originally proposed.
We initialize our CNN using the pre-trained model and fine-tune the network on the corresponding
training set.

Again for fair comparison,
for the deep CNN approach SFHC,  we replace its network structure (convolution-pooling, fully-connected layers)
with the VGG-16 model and end-to-end train the network based on the triplet hinge loss used in the original paper.
We  implement SFHC using \textit{Theano}~\cite{bastien2012theano} and train the model
using two GeForce GTX Titan X.
The triplet samples are randomly generated in the course of training, following \cite{lai2015simultaneous}.

For the NUS-WIDE dataset, we construct two comparison settings, setting-1 and setting-2.
For setting-1, following the previous work %
\cite{lai2015simultaneous, liu2011hashing}, we consider the 21 most frequent tags and the similarity
is defined based on whether two images share at least one common tag.
For setting-2, we use the similarity precision evaluation metric to evaluate pairwise and triplet performance.
As in \cite{wang2014learning}, similarity precision is defined as the \% of triplets being correctly ranked.

Given a triplet image set $({{\bf{x}}_i},{{\bf{x}}_j},{{\bf{x}}_k})$, where $s({{\bf{x}}_i},{{\bf{x}}_j}) > s({{\bf{x}}_i},{{\bf{x}}_k})$.
We assume ${{\bf{x}}_i}$ as the query, if the rank of ${{\bf{x}}_j}$ is higher than ${{\bf{x}}_k}$, then we say triplet is correctly ranked.
We first randomly sample 1000 probe images from all the data sharing the selected 21 attributes in setting-1.
Then we obtain a ranking list for each probe image according to how many attributes it shares
with the data and randomly generate 50 triplets per probe image according to the ranking list to form the test set.
For the triplet-based methods, the sampled training data is the same as in setting-1.
For the compared pairwise-based methods, we directly use the hash functions learned in setting-1
since semantic ranking information cannot be incorporated into the pairwise-based inference pipeline.
For CIFAR-10 and NUS-WIDE setting-1, we use the same experimental setting as described in ~\cite{lai2015simultaneous}.

\begin{figure*}[tbp]
	\centering
	\resizebox{0.95\linewidth}{!}
	{
		\begin{tabular}{c@{}c@{}c}
			\includegraphics{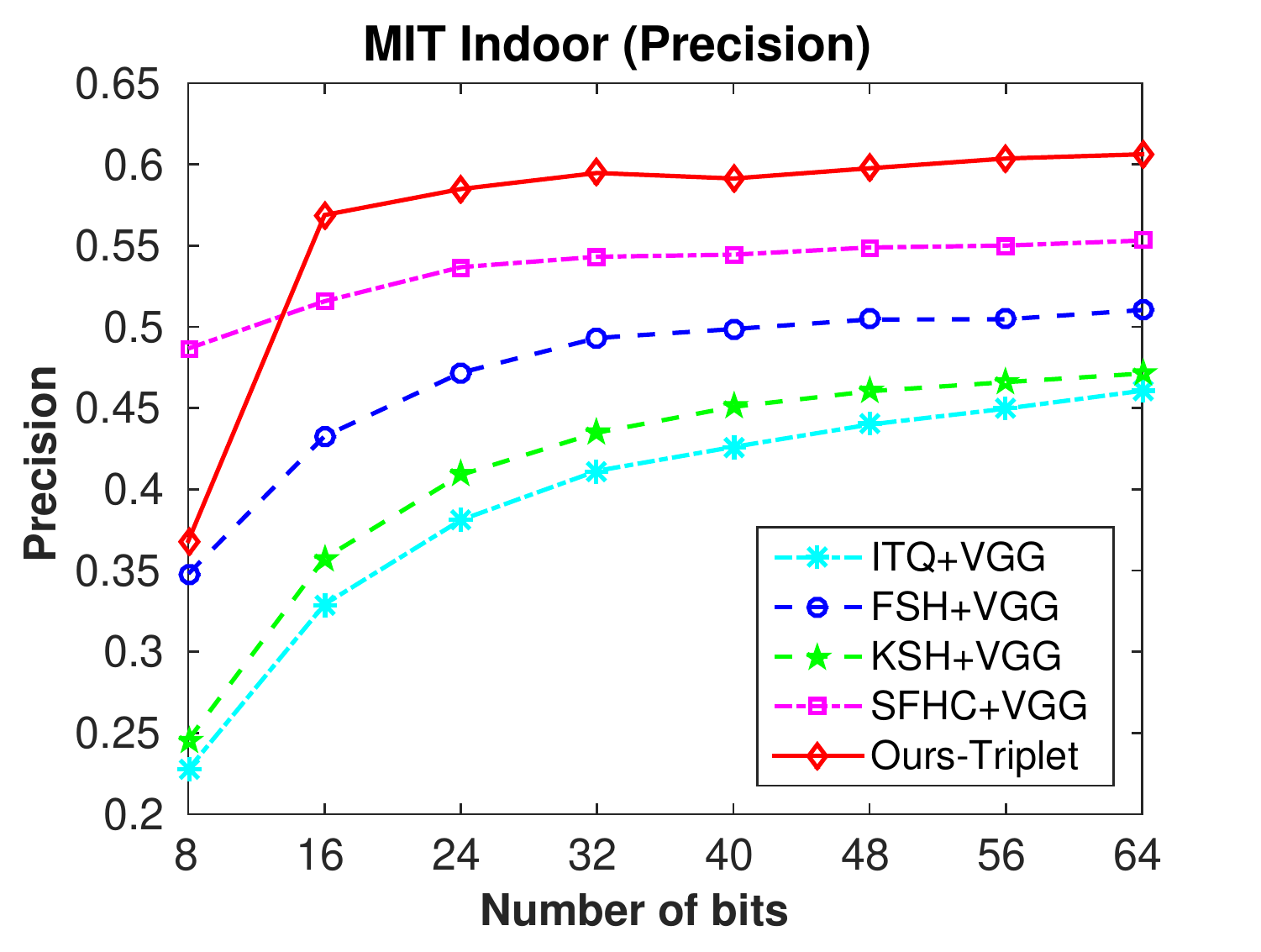}
			&
			\includegraphics{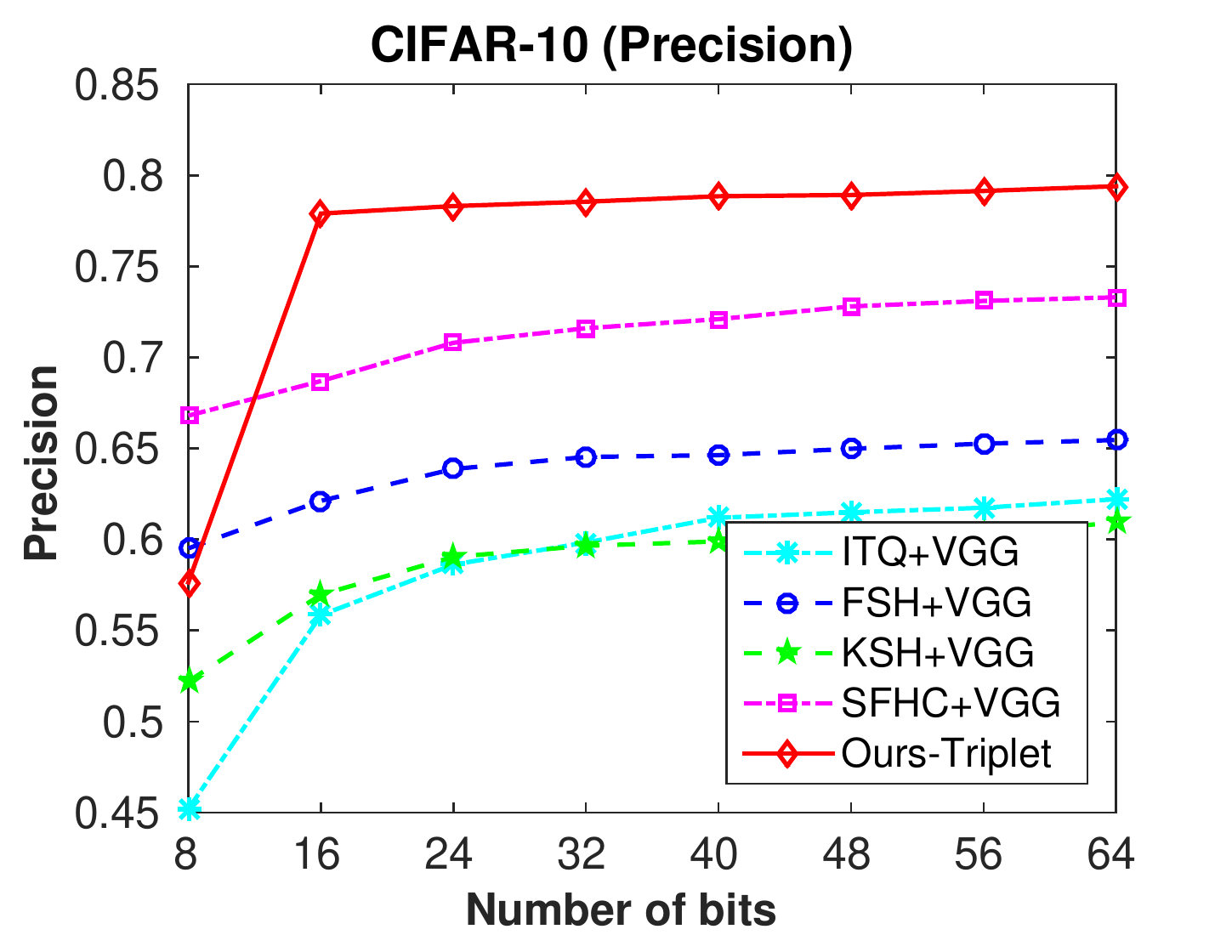}
			&
			\includegraphics{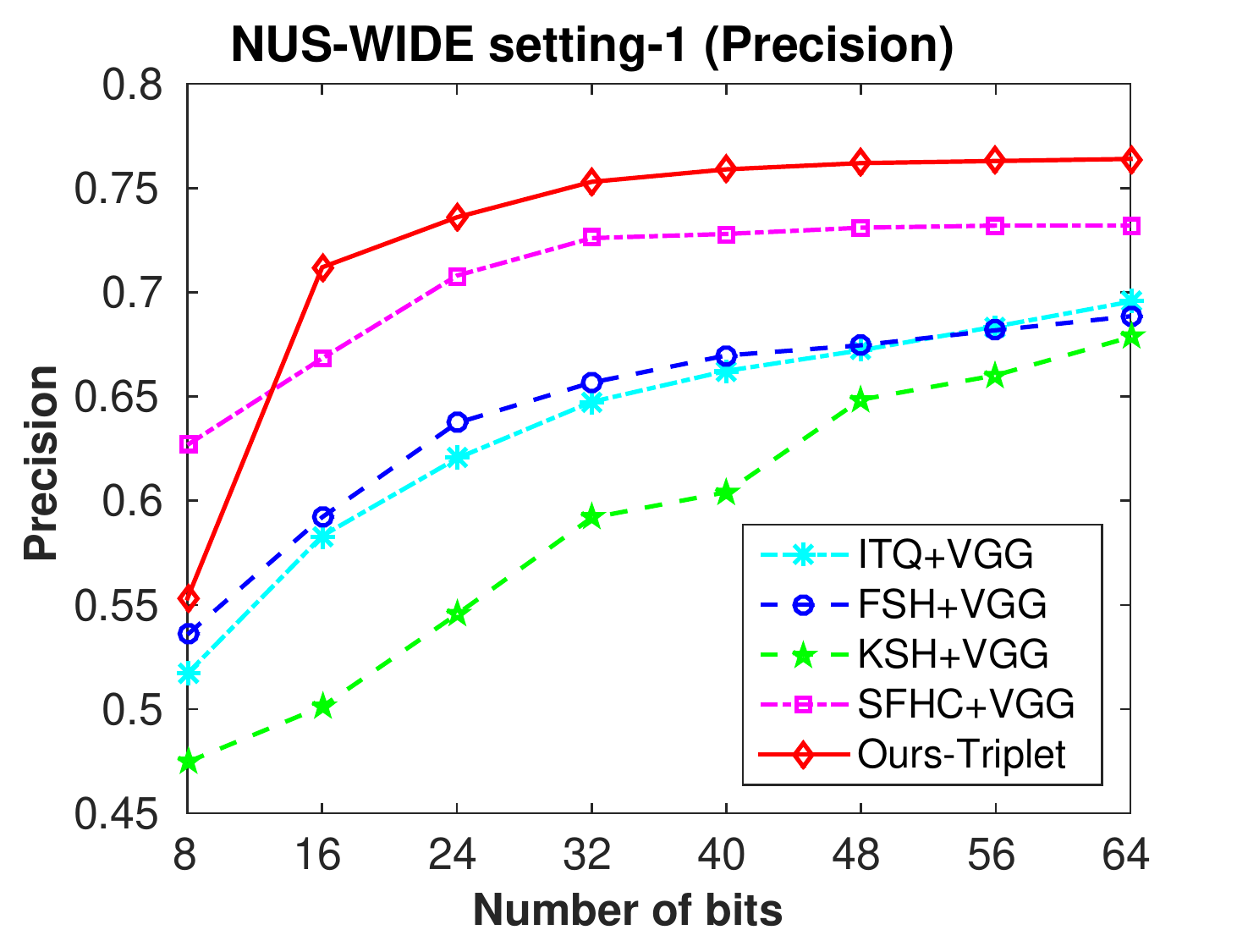}
			\\

		\end{tabular}
	}
		\caption{The precision curves on three datasets.
		We compare several state-of-the-art algorithms including ITQ~\cite{gong2013iterative}, KSH~\cite{ liu2012supervised}, FSH~\cite{lin2014fast} with features extracted from VGG-16 model which is fine-tuned on the corresponding training set and SHFC~\cite{lai2015simultaneous} which is implemented using the VGG-16 network structure. }
	\label{fig:precision}

\end{figure*}

\begin{figure*}[tbp]

	\centering
	\resizebox{0.95\linewidth}{!}
	{
		\begin{tabular}{c@{}c@{}c}
			\includegraphics{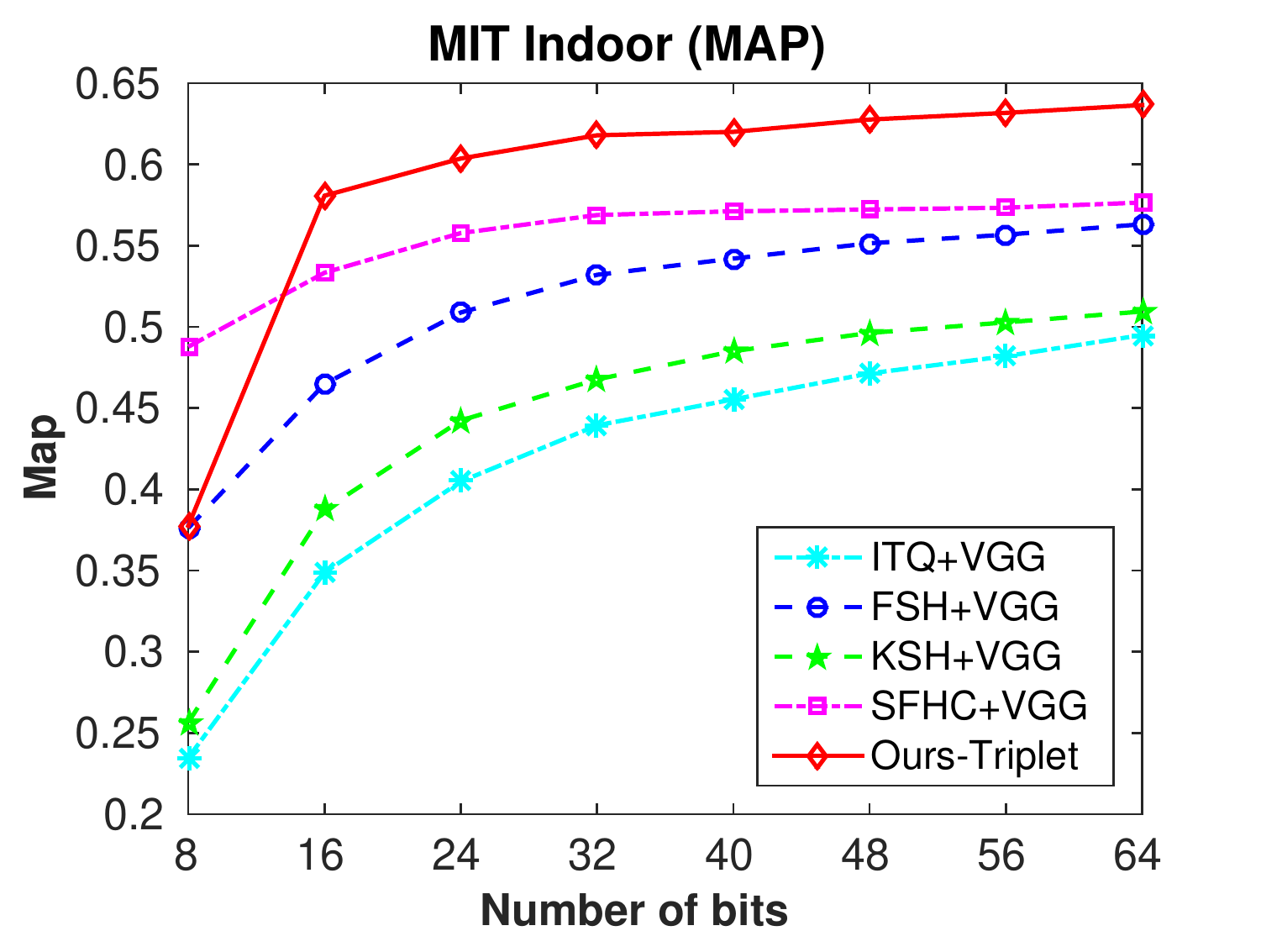}
			&
			\includegraphics{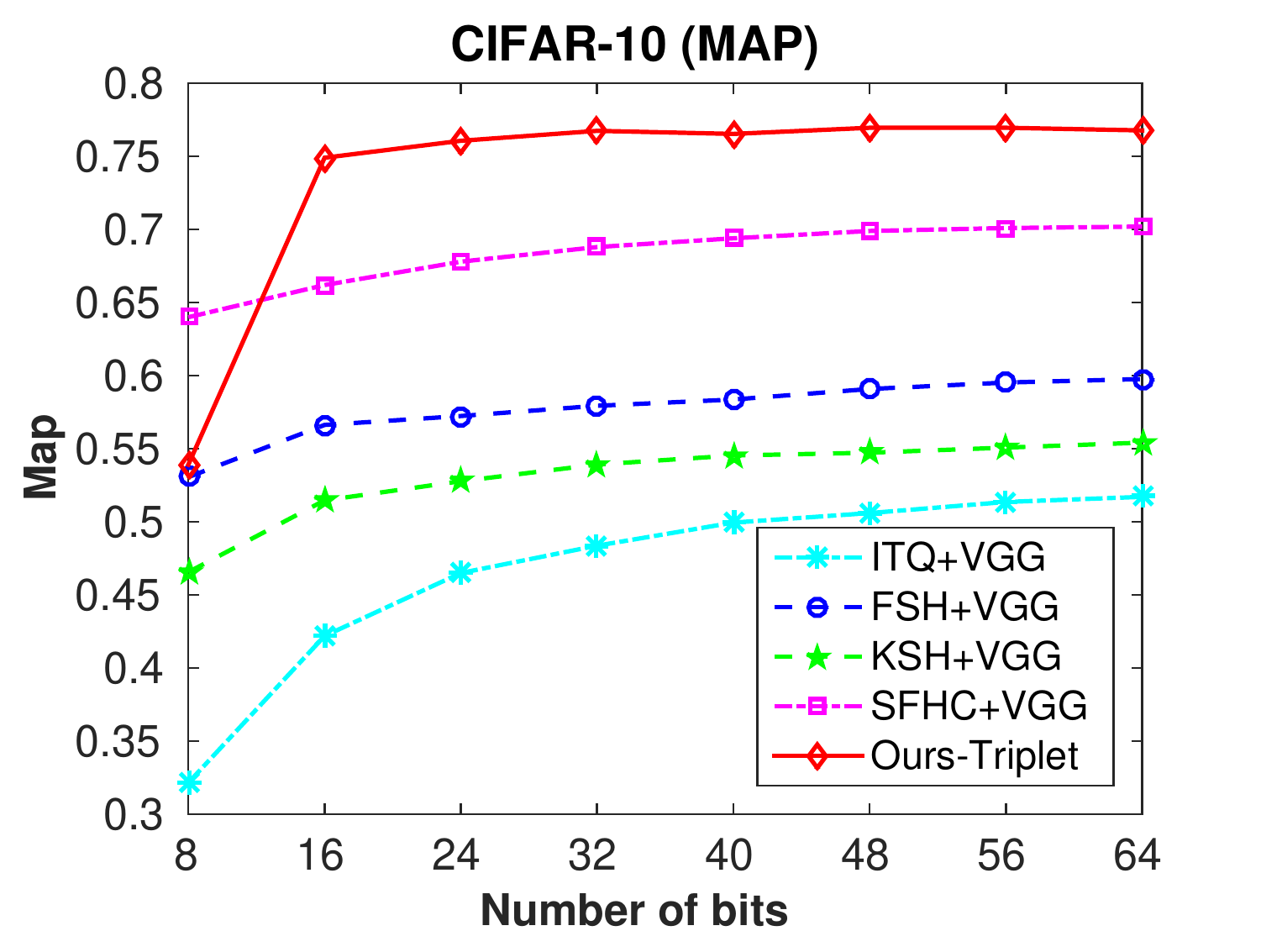}
			&
			\includegraphics{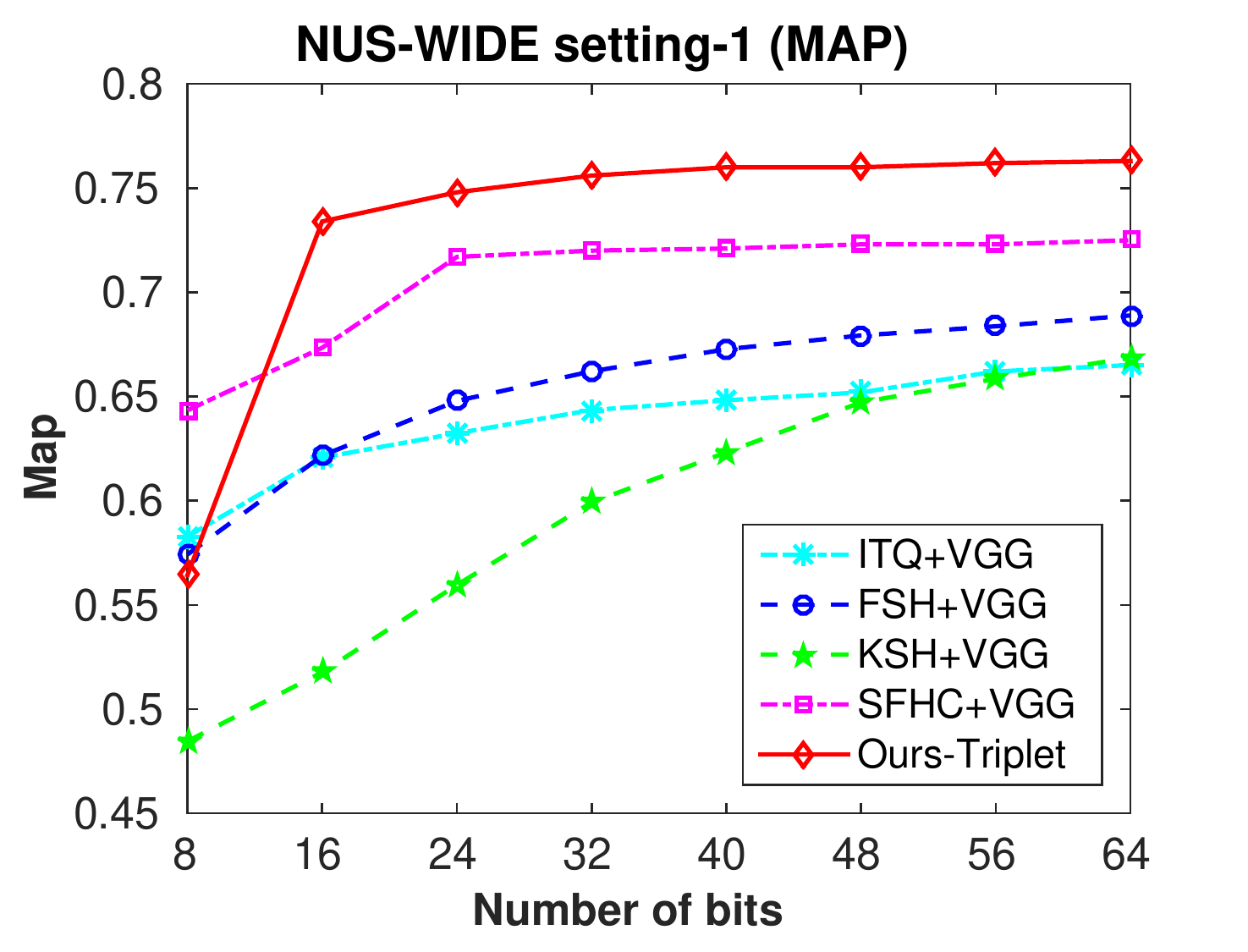}
			\\

		\end{tabular}
	}
		\caption{The mean average precision curves on three datasets.
		Settings are the same as in Figure \ref{fig:precision}.}
	\label{fig:map}

\end{figure*}

We use two evaluation metrics: Mean Average Precision (MAP) and the precision of the top-K retrieved examples (Precision),
where K is set to 100 in CIFAR-10 and NUS-WIDE setting-1 and set to 80 in MIT Indoor dataset.
For NUS-WIDE setting-1, we calculate the MAP values within the top 5000 returned neighbors.
The results are represented in Figure~\ref{fig:precision} and Figure~\ref{fig:map}.

\subsection{Implementation details}

We implement the network training based on the CNN toolbox \textit{Theano}.
Training is done on a standard desktop with a GeForce GTX TITAN X with 12GB memory.
In all experiments, we set the mini-batch size for gradient descent to 50,
momentum  0.9, weight decay  0.0005 and dropout rate  0.5 on the fully connected layer to avoid over-fitting.
The number of binary codes per group is set to 8.

\subsection{Analysis of retrieval results}

\begin{figure}[tbp]
	\caption{The similarity precision curves on NUS-WIDE setting-2.}
	\label{fig:NUS_2}
	\centering
	\resizebox{0.98\linewidth}{!}
	{\begin{tabular}{c}
			\includegraphics{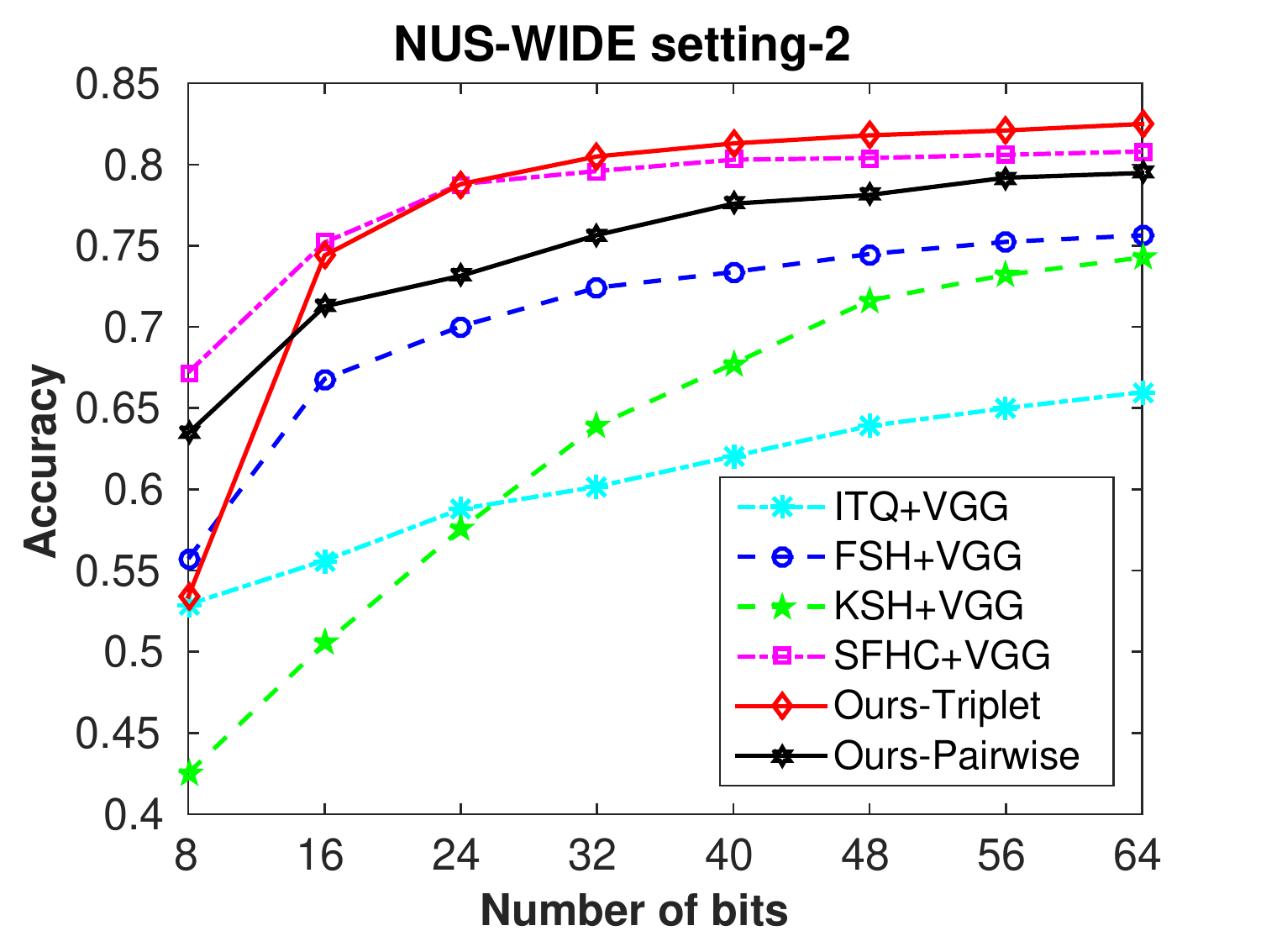}
		\end{tabular}
	}
\end{figure}

\begin{figure}[tbp]
	\caption{Evaluation of the inference performance on three datasets.}
	\label{fig:code}
	\centering
	\resizebox{0.98\linewidth}{!}
	{\begin{tabular}{c}
			\includegraphics{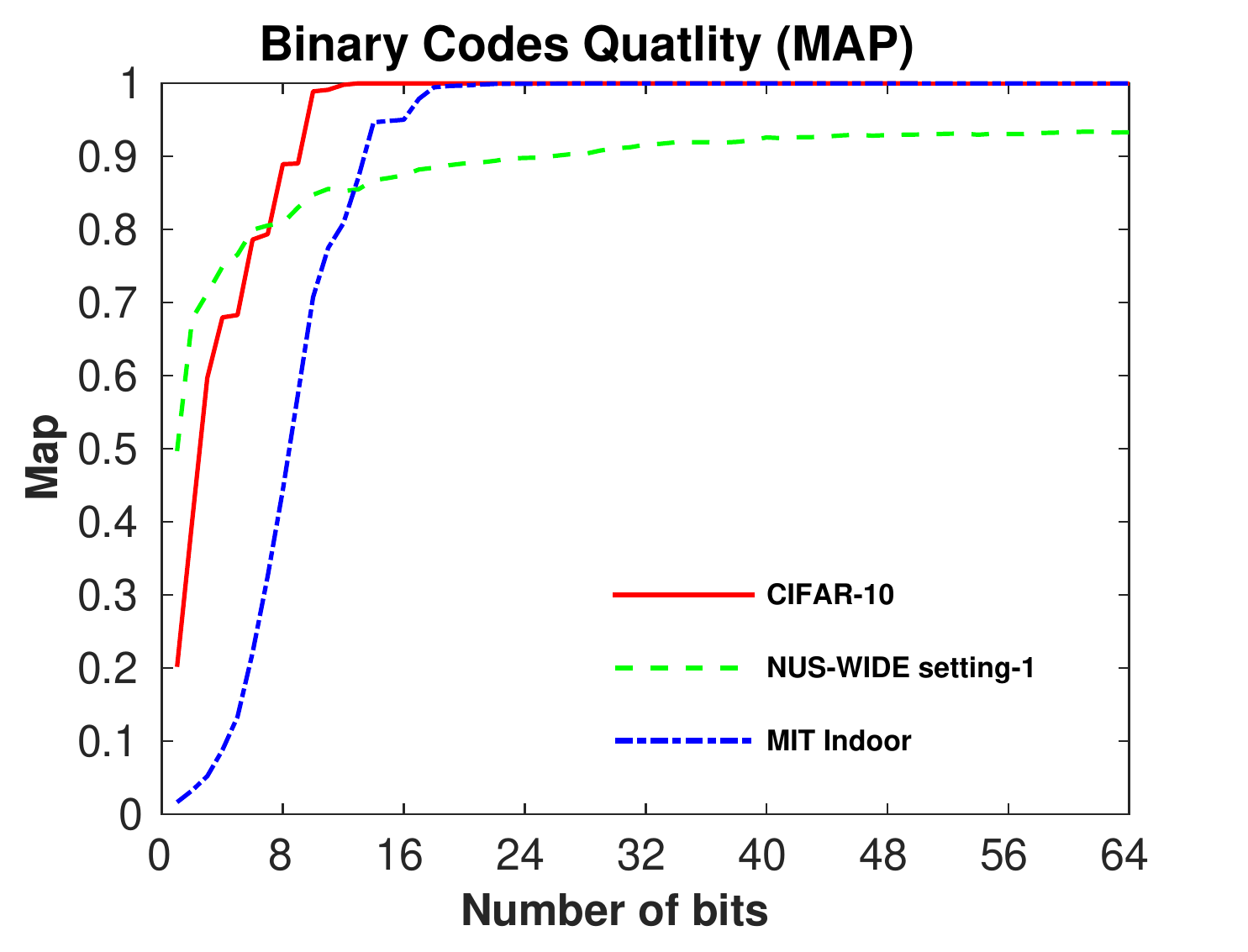}
		\end{tabular}
	}
\end{figure}

On all  the three datasets,  our
proposed method shows superior performance in terms of MAP and precision evaluation metrics
against the most related work SFHC (deep CNN) and FSH (two-step hashing with boosted trees).
As expected, the training speed of our method is much faster than SFHC, and the result is summarized in Table~\ref{tab:time}.
Rather than simply end-to-end learn the hash functions,
our method incorporates hash functions learning with a collaborative inference step,
where the image representation learning and hash coding can benefit each other through this feedback scheme.

Compared to FSH, the results demonstrate the effectiveness of incorporating relative similarity information as supervision.
Note that FSH is based on pairwise information while ours uses triplet based ranking information to learn hash codes.
The triplet loss may be better for retrieval tasks because it is directly linked to retrieval measure such as the AUC score.
The pairwise loss used by FSH  encourages all images in one category  to be projected
onto a single point in the Hamming space.
The triplet loss maximizes a margin between each
pair of same-category images and images from different categories.
As argued in \cite{schroff2015facenet, LMNN},  this may enable images belonging to the same category  to reside on a manifold;
and at the same time to maintain a distance from other categories.

\begin{table*}[h!]
	\caption{Training time of the proposed method and the method SFHC~\cite{lai2015simultaneous} on three datasets.
		In terms of training time, our method is significantly faster than SFHC.}
	\centering
  \resizebox{.7\linewidth}{!}
	{
		\begin{tabular}{c|c c c |c}
			\hline
			\multirow{2}{*}{Method}&\multicolumn{3}{c|}{Training Time (hours)}&\multirow{2}{*}{Number of GPUs}\\
			&MIT Indoor&CIFAR-10&NUS-WIDE setting-1\\\hline
			Ours-Triplet&18&15&32&1\\\hline
			SFHC&186&174&365&2\\\hline
		\end{tabular}
	}
	\label{tab:time}
\end{table*}

\subsection{Triplet vs.\ pairwise}

From the results shown in Figure~\ref{fig:NUS_2}, we can clearly
observe the superiority of triplet-based methods on the ranking based evaluation metric. Thanks to the high quality  binary codes and the strong fitting capability of our deep model,
our proposed method provides much better performance than pairwise methods by a large margin.

Since the two triplet-based methods (Ours-Triplet and SFHC) simultaneously learn feature representations
and hash codes while considering the semantic ranking information, they are
more likely to learn hash functions that are tailored for the ranking-based retrieval metric than the pairwise-based methods
(Ours-pairwise and FSH).

\subsection{Evaluation of binary codes quality}

\begin{table*}[htp]
	\caption{Face search accuracies under the IJB-A protocol.
		Results for GOTS and OpenBR are quoted from~\cite{klare2015pushing}.
		Results are reported as the average $ \pm $ standard deviation over the 10-fold cross validation sets specified in the IJB-A protocol.  }
	\centering
	\scalebox{0.86}
	{
		\begin{tabular}{c|c|c|c|c}
			\hline
			\multirow{2}{*}{Algorithm}& \multicolumn{2}{c|}{CMC (closed-set search)} &\multicolumn{2}{c}{FNIR @ FPIR (open-set search)} \\\cline{2-5}& Rank-1&Rank-5&0.1&0.01\\\hline
			GORS&$0.443 \pm 0.021$&$0.595 \pm 0.020$&$0.765 \pm 0.033$&$0.953 \pm 0.024$\\\hline
			OpenBR&$0.246 \pm 0.011$&$0.375 \pm 0.008$&$0.851 \pm 0.028$&$0.934 \pm 0.017$\\\hline
			Deep Face Search\cite{wang2015face}&$0.820 \pm 0.024$&$0.929 \pm 0.013$&$0.387 \pm 0.032$&$0.617 \pm 0.063$\\\hline
			Proposed Method&$\bf{0.831 \pm 0.020}$&$\bf{0.937 \pm 0.015}$&$\bf{0.369 \pm 0.028}$&$\bf{0.598 \pm 0.048}$\\\hline

		\end{tabular}}
		\label{tab:face}
	\end{table*}

	We evaluate the binary codes quality on CIFAR-10, MIT Indoor and NUS-WIDE setting-1 datasets (see Figure~\ref{fig:code}).
	To evaluate the effectiveness of the binary codes  inference pipeline,
	we infer 64 binary bits without learning the deep hash functions.
	Then the training database is used as both the probe set and the gallery set for evaluating the inference performance.
	For the three datasets, we calculate the MAP values within the returned neighbors.
	We can observe that for CIFAR-10, the binary codes converge very fast at around 10-th bits.
	MIT Indoor dataset converges slightly slower due to the fact that it has more classes.
	The binary codes can still perfectly separate all the training samples from different classes.
	This is because the relations between training points are very simple due to the multi-class similarity relationships.
	In contrast, due to the complicated relationships between the multi-label training samples,
	the accuracy of NUS-WIDE setting-1 keeps improving up to 64 bits and is lower than those multi-class datasets.
	We can see that the code quality is directly proportional to the final retrieval performance.
	This makes sense since the deep hash functions are learned to fit the binary codes,
	so the performance of the inference pipeline has a  direct impact on the quality of the learned deep hash functions.

	\subsection{Face retrieval}

	We implement the face search application as follows.
	\noindent \textit{Data preprocessing.} The preprocessing pipeline is:
	1) detect the face region using the robust face detector \cite{mathias2014face}
	and find 68 face landmarks using the (state-of-the-art) face alignment algorithm \cite{xiong2013supervised};
	2) select the middle landmark between two eyes and the middle landmark of the mouth
	as alignment-anchor points, and align/scale the face image such that distance between the landmarks is 40 pixels;
	3) finally we crop a $160 \times 160$ region around the mid-point of the two landmarks in (2).

\begin{table*}[htp]
	\caption{Face search accuracies of the proposed method under the IJB-A protocol using different bits per group.  }
	\centering
	\scalebox{0.86}
	{
		\begin{tabular}{c|c|c|c|c}
			\hline
			\multirow{2}{*}{Group length}& \multicolumn{2}{c|}{CMC (closed-set search)} &\multicolumn{2}{c}{FNIR @ FPIR (open-set search)} \\\cline{2-5}& Rank-1&Rank-5&0.1&0.01\\\hline
			8 bits&$\bf{0.831\pm 0.020}$&$\bf{0.937 \pm 0.015}$&$\bf{0.369 \pm 0.028}$ &$\bf{0.598 \pm 0.048}$\\\hline
			32 bits&$0.818 \pm 0.023$&$0.920 \pm 0.016$&$0.385 \pm 0.030$&$0.612 \pm 0.052$ \\\hline
			64 bits&$0.793 \pm 0.024$&$0.908 \pm 0.018$&$0.398 \pm 0.036$ &$0.627 \pm 0.061$ \\\hline
			128 bits&$0.778 \pm 0.023$&$0.889 \pm 0.020$&$0.415 \pm 0.035$&${\rm{0.645}} \pm 0.058$\\\hline

		\end{tabular}}
		\label{tab:group}
	\end{table*}

	\noindent \textit{Supervised pre-training}.
	We pre-train the VGG-16 \cite{simonyan2014very} network (using \textit{Caffe} \cite{jia2014caffe}) to classify all the 10575 subjects in the CASIA dataset~\cite{yi2014learning}.  This dataset has 494414 images of the 10575 subjects, and we double the number of training examples by horiozontal mirroring, making the feature representation more robust to pose variation.

	We test the pre-trained model's discriminative power on the LFW verification data as follows.
	We use the last 4096-dimensional fully-connected layer as the feature representation
	and then use PCA to compress it into a 160-dimensional feature vector.
	Then CNN features are centered and normalized for evaluation.
	Under the standard LFW \cite{huang2007labeled} face verification protocol, for a single network using only cosine similarity, we achieve an accuracy of $\bf{97.03}\%  \pm \bf{0.98}\%$.
	Using the joint Bayesian method~\cite{chen2012bayesian} for face verification,
	we achieve an accuracy of $\bf{98.18}\%  \pm \bf{0.96}\%$.

	Despite using only publicly available training data and one single network,
	the performance of this model is competitive with state-of-the-art~\cite{schroff2015facenet,taigman2014deepface, yi2014learning, sun2015deepid3}.

	\noindent \textit{Face search.}
	We then use the above pre-trained CNN model  to initialize the deep CNN that models the hash functions of our proposed hashing method.
	We test the face search performance on the
	IARPA Janus Benchmark-A (IJB-A) dataset \cite{klare2015pushing} which contains 500 subjects with a total of 25,813 face images.
	This dataset contains many challenging face images and defines both verification and search protocols.
	The search task (1:$N$ search) is defined in terms of comparisons between templates consisting of several face images,
	rather than single face images.
	For the search protocol, which evaluates both closed-set and open-set search performance,
	10-fold cross validation sets are defined based on both the probe and gallery sets consisting of templates.
	Given an image from the IJB-A dataset, we first detect and align the face following the data preprocessing pipeline.
	After processing, the final training set consists approximately 1 million faces and 1 billion randomly sampled triplets.
	Clearly,  such a large-scale training dataset may render  most existing triplet-based hashing methods computationally intractable.
	The deep hash functions are learned based on the proposed two-step hashing framework.
	After the deep hash functions are learned, we  generate $ 128 $ bits hash codes for each input face image for fast face retrieval.
	The definitions of CMC, FNIR and FPIR are explained in \cite{wang2015face, klare2015pushing}.
	The results of the proposed method along with the compared algorithms are reported in Table~\ref{tab:face}.
	In \cite{wang2015face}, a face is represented by the combined features extracted by 6 deep models.
	However, in our paper, 128 bits binary codes are directed extracted by a single deep model for face representation which enjoys both faster searching speed and less storage space.
	Also, although using the same training database, the searching accuracy on two protocols both
	demonstrate the effectiveness of our hashing framework.

\subsection{Evaluation of the incremental learning}

	We evaluate different group lengths used in the incremental learning to prove the effectiveness of such an optimization strategy.
	We implement the experiments on the face retrieval task as described above since there are sufficient training examples and faces are difficult for the deep architecture to fit because of the relatively weak discriminative information they share.
	The results are reported in Table~\ref{tab:group}.
	From the results, we clearly see that smaller group length corresponds to better search accuracies, demonstrating our assertion that incremental optimization helps in terms of code quality and the final performance.

	\section{Conclusion}

	In this paper, we develop a general  supervised hashing method with triplet ranking loss for large-scale image retrieval.
	Instead of directly training on the extremely large amount of triplet samples,
	we formulate learning of the deep hash functions as a multi-label classification problem, which allows us to learn deep hash functions  orders of magnitude faster than the previous triplet based hashing methods
	in terms of training speed.
	The deep hash functions are learned in an incremental scheme, where the inferred binary codes
	are used to learn image representations and the learned hash functions can give feedback for boosting the quality of binary codes.
	Experiments demonstrate that the superiority of the proposed method over other state-of-the-art hashing methods.

{\small
\bibliographystyle{ieee}
\bibliography{draft}
}

\end{document}